%% file: main.tex
\documentclass[conference]{IEEEtran}
\IEEEoverridecommandlockouts

\usepackage{cite}
\usepackage{amsmath,amssymb,amsfonts}
\usepackage{graphicx}
\usepackage{textcomp}
\usepackage{xcolor}
\usepackage{mathtools}
\usepackage{amsthm}
\usepackage{cutwin}
\usepackage{bm}
\usepackage{enumitem}
\usepackage{amsmath}
\usepackage{multirow}
\usepackage{color}
\usepackage{amsfonts}
\usepackage{nicefrac}
\usepackage{bbding}
\usepackage{pifont}
\usepackage{array} 
\usepackage{colortbl}
\usepackage{rotating}
\usepackage{adjustbox}
\usepackage{makecell}
\usepackage{tcolorbox}
\usepackage{fbox}
\usepackage{algpseudocode}

\usepackage{booktabs}
\usepackage{amssymb}
\usepackage{multirow}
\usepackage{array}
\usepackage{arydshln}
\usepackage[utf8]{inputenc}

\usepackage{booktabs}
\usepackage{multirow}
\usepackage{arydshln}
\usepackage{tabularx}
\usepackage{pifont}

\definecolor{cvprblue}{rgb}{0.21,0.49,0.74}
\definecolor{sgreen}{RGB}{30, 150, 30}
\definecolor{mycolor_blue}{HTML}{E7EFFA}
\definecolor{mycolor_green}{HTML}{E6F8E0}
\definecolor{mycolor_gray}{HTML}{ECECEC}
\definecolor{pearDark}{HTML}{2980B9}
\definecolor{textcolor1}{rgb}{0.25,0.5,0.5}
\definecolor{textcolor2}{rgb}{0.7,0.25,0.25}
\definecolor{linkc}{rgb}{0, 0.44, 0.74}
\definecolor{eqc}{rgb}{1, 0, 0}
\definecolor{myy}{RGB}{126,95,0}
\definecolor{mygray}{gray}{.9}
\definecolor{bblue}{RGB}{30,80,120}
\definecolor{mygray1}{gray}{.7}
\definecolor{ggray}{RGB}{127,127,127}
\definecolor{mygreen}{RGB}{93,174,86}
\definecolor{citecolor}{HTML}{229954}
\definecolor{light_green}{HTML}{F5FFFA}
\definecolor{LightCyan}{rgb}{0.88,1,1}
\definecolor{scolor}{RGB}{111,168,220}
\definecolor{hcolor}{RGB}{111,176,81}
\definecolor{ocolor}{RGB}{224,103,102}
\definecolor{wcolor}{RGB}{246,178,107}

\usepackage[pagebackref,breaklinks,colorlinks,allcolors=cvprblue]{hyperref}

\def\BibTeX{{\rm B\kern-.05em{\sc i\kern-.025em b}\kern-.08em
    T\kern-.1667em\lower.7ex\hbox{E}\kern-.125emX}}

\begin{document}
\title{Disentangling Hardness from Noise: An Uncertainty-Driven Model-Agnostic Framework \\ for Long-Tailed Remote Sensing Classification}

\author{
    \IEEEauthorblockN{
        \textbf{Chi Ding}$^{1*}$, 
        \textbf{Junxiao Xue}$^{1*}$, 
        \textbf{Xinyi Yin}$^{2}$, 
        \textbf{Shi Chen}$^{3}$, 
        \textbf{Yunyun Shi}$^{3}$ \\
        \textbf{Yiduo Wang}$^{2}$, 
        \textbf{Fengjian Xue}$^{3}$, 
        \textbf{Xuecheng Wu}$^{3\dagger}$
    }
    \IEEEauthorblockA{
        $^{1}$Research Center for Space Computing System, Zhejiang Lab, Hangzhou, China \\
        $^{2}$School of Cyber Science and Engineering, Zhengzhou University, Zhengzhou, China \\
        $^{3}$School of Computer Science and Technology, Xi'an Jiaotong University, Xi'an, China
    }
    \IEEEauthorblockA{
        \texttt{ xuejx@zhejianglab.cn, xuecwu@gmail.com}
    }
    \thanks{$^*$Equal contribution.}
    \thanks{$^\dagger$Corresponding author.}
}

\maketitle

\input{Sec/0_abs}


\input{Sec/1_intro}
\input{Sec/2_related}
\input{Sec/3_method}
\input{Sec/4_expers}
\input{Sec/5_conclusions}

\bibliographystyle{IEEEtran}
\bibliography{main}

\end{document}

%% file: Sec/0_abs.tex
\begin{abstract}
Long-Tailed distributions are pervasive in remote sensing due to the inherently imbalanced occurrence of grounded objects. However, a critical challenge remains largely overlooked, \textit{i.e.}, disentangling hard tail data samples from noisy ambiguous ones. Conventional methods often indiscriminately emphasize all low-confidence samples, leading to overfitting on noisy data. To bridge this gap, building upon Evidential Deep Learning, we propose a model-agnostic uncertainty-aware framework termed DUAL, which dynamically disentangles prediction uncertainty into Epistemic Uncertainty (EU) and Aleatoric Uncertainty (AU). Specifically, we introduce EU as an indicator of sample scarcity to guide a reweighting strategy for hard-to-learn tail samples, while leveraging AU to quantify data ambiguity, employing an adaptive label smoothing mechanism to suppress the impact of noise. Extensive experiments on multiple datasets across various backbones demonstrate the effectiveness and generalization of our framework, surpassing strong baselines such as TGN and SADE. Ablation studies provide further insights into the crucial choices of our design.
\end{abstract}

%% file: Sec/1_intro.tex
\section{Introduction}
\label{sec:intro}

In the remote sensing image scenarios \cite{xia2018dota,li2020dior}, land cover categories typically exhibit a significant imbalanced distribution, commonly referred to as a long-tailed distribution: a small number of frequent categories possess a large number of samples, while the majority of rare categories have only a limited number of samples available for learning. This distribution poses substantial challenges to the representational capacity of deep learning models \cite{zhang2023deepsurvey}, where models tend to overfit to head classes and perform poorly on tail classes.

In recent years, research on long-tailed distribution has mainly focused on methods such as resampling \cite{cui2019class}, loss reweighting \cite{LDAM}, and logit adjustment \cite{chen2025robust_logit}. Resampling balances the class distribution by oversampling tail classes or undersampling head classes. Loss reweighting approaches, such as Class-Balanced Loss \cite{cui2019class} and Focal Loss \cite{ross2017focal}, adjust loss weights according to class frequency or sample difficulty to improve tail-class performance.
While they are effective to some extent, a critical challenge in remote sensing remains largely overlooked: the distinction between hard-to-learn tail samples and noisy ambiguous samples. Unlike natural images, remote sensing images often suffer from variations in sensor resolution, cloud occlusion, overlapping land cover, and changes in lighting conditions, introducing inherent noise or ambiguity. 

Conventional methods typically rely solely on class frequency or prediction logits to evaluate the importance of the sample. Consequently, they indiscriminately emphasize all hard-to-learn samples, leading to overfitting on noisy data rather than mitigating their negative influence.

As a result, the core research challenge arises: \textit{How can we encourage the model to learn from rare samples while simultaneously suppressing the impact of noisy data?}
To address this core challenge, we introduce uncertainty estimation to better disentangle the rare samples and noisy data. Uncertainty is typically categorized into two types: Epistemic Uncertainty (EU) and Aleatoric Uncertainty (AU) \cite{zhang2023deepsurvey}. EU reflects the model's lack of knowledge, often arising from regions in the input space where the model has not been sufficiently trained or cannot make confident predictions. In contrast, AU captures the inherent noise or ambiguity in the data, which cannot be reduced simply by collecting more data.

The two types of uncertainty precisely match two key issues in long‑tailed remote sensing: insufficient learning of tail classes by the model (EU) and quality degradation or semantic ambiguity in a subset of samples (AU). Therefore, compared with methods that rely solely on class frequencies or logits, uncertainty provides a more fine-grained training signal. On the one hand, EU serves as an indicator of samples that are currently under-learned by the model, helping identify which instances deserve prioritized training. On the other hand, AU evaluates the quality of each sample, enabling dynamic adjustment of the supervision strength to avoid overfitting noisy or semantically ambiguous data.

Building on this insight, we propose a dual uncertainty-aware long-tailed learning framework, termed \textbf{DUAL}. Firstly, we adopt Evidential Deep Learning (EDL) \cite{sensoy2018evidential} to dynamically model uncertainty and disentangle EU from AU during training. We then perform EU-Based sample reweighting, reallocating weights according to the model’s current epistemic state to emphasize samples that require more learning, and introduce an AU-guided dynamic label smoothing approach that adapts the supervision strength based on the estimated aleatoric uncertainty.

In summary, the main contributions of this paper are three-fold:
\begin{itemize}
    \item \textbf{A novel uncertainty-aware framework to disentangle hardness from noise.} We identify the critical limitation of existing methods in disentangling hard tail samples from noisy ambiguous ones. 
    To address this, we propose \textbf{DUAL}, a model-agnostic framework based on EDL,  which dynamically disentangles prediction uncertainty into EU and AU to identify the source of low confidence.
    
    \item \textbf{A dynamically guided training framework driven by uncertainty.} We design an uncertainty-guided mechanism to handle tail and noisy samples in \textbf{DUAL}. Specifically, \textbf{DUAL} utilizes EU to indicate sample scarcity for reweighting hard tail samples, while leveraging AU to measure data ambiguity for adaptive label smoothing to suppress noise.
    
    \item \textbf{Extensive validation on multiple long-tailed remote sensing benchmarks.} Experiments demonstrate that \textbf{DUAL} consistently improves overall accuracy and significantly boosts tail-class performance, confirming its effectiveness and practicality.
\end{itemize}

%% file: Sec/2_related.tex
\section{Related Work}
\label{sec:Related Work}
\subsection{Long-tailed Learning}
Long-tailed data is a common challenge in many real-world scenarios including remote sensing image analysis, where the distribution of classes is heavily imbalanced. In these datasets, a few classes referred to as head classes dominate the majority of samples, while many other classes referred to as tail classes are underrepresented. This imbalance poses significant challenges for machine learning models, as standard training approaches tend to overfit head classes while underperforming on tail classes. 
Re-sampling \cite{cui2019class} and class-sensitive learning \cite{ross2017focal} are the dominant methods to deal with long-tailed data. Resampling balances the data distribution by adjusting the sampling weights of the samples, and class-sensitive learning deals with the imbalance of the data distribution by adjusting the loss function of the model. 
However, these methods typically rely solely on class frequency or prediction logits. Consequently, they tend to indiscriminately emphasize all tail or hard-to-learn samples, ignoring the inherent quality issues within the data. By failing to identify samples that are unsuitable for training (\textit{e.g.}, due to noise or ambiguity), these approaches often lead to overfitting rather than mitigating the negative influence of low-quality samples.

\subsection{Uncertainty Estimation}
In recent years, deep neural networks have achieved remarkable success across various domains~\cite{wu2025avf,wu2025scalable,wu2025vic,xue2024affective}. However, as these models are increasingly deployed in real-world applications, the reliability of their predictions has become a critical concern. Uncertainty in deep learning is typically categorized into model uncertainty (epistemic), arising from knowledge gaps due to limited data, and data uncertainty (aleatoric), caused by inherent noise. Early approaches \cite{zhang2023deepsurvey} to estimate these uncertainties include Bayesian Neural Networks (BNNs), Deep Ensembles, and Monte Carlo (MC) Dropout. However, these methods often incur high computational costs due to multiple forward passes (Ensembles, MC Dropout) or suffer from convergence difficulties (BNNs), limiting their practicality in large-scale remote sensing. In contrast, EDL \cite{sensoy2018evidential} offers a deterministic and efficient alternative. By modeling the predictive distribution as a Dirichlet distribution, EDL enables the simultaneous quantification of prediction, epistemic uncertainty, and aleatoric uncertainty within a single forward pass. Crucially, EDL can estimate uncertainty dynamically during training without altering the backbone architecture or requiring expensive sampling, making it highly suitable for optimizing the model training process.

\begin{figure*}[!t]
    \centering
    \includegraphics[width=1.0\linewidth]{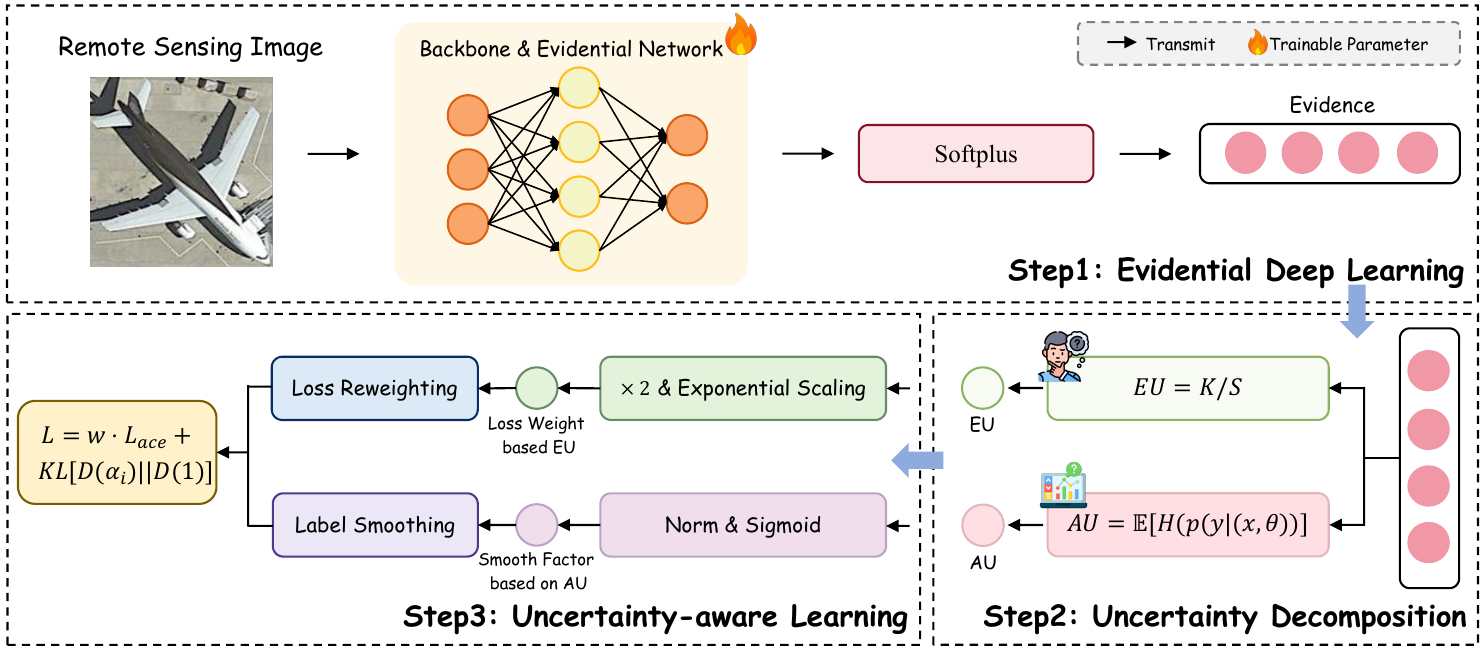}
    \caption{The overview of our proposed DUAL framework. The pipeline consists of three stages: (1) Evidential Deep Learning, which predicts class-level evidence from the backbone; (2) Uncertainty Decomposition, which decomposes prediction uncertainty into EU and AU; and (3) Uncertainty-aware Learning, where EU serves as an indicator of sample scarcity to reweight hard tail samples, while AU quantifies data ambiguity to guide adaptive label smoothing for noise suppression.}
    \label{fig:workflow}
\end{figure*}

%% file: Sec/3_method.tex
\section{Methodology}
\label{sec:Mthods}
As illustrated in Figure \ref{fig:workflow}, \textbf{DUAL} consists of three key components: (1) uncertainty estimation by EDL; (2) disentangling EU and AU from model predictions; (3) EU-Based sample reweighting to address insufficient learning of tail classes, and AU-Based dynamic label smoothing to reduce the impact of ambiguous samples.

\subsection{Evidential Deep Learning}

In EDL, the parameters of the Dirichlet distribution need to be determined to evaluate uncertainty. Evidence refers to the indicators obtained from the inputs to support categorization, and is closely related to the parameters of the Dirichlet distribution. According to Dempster-Shafer Evidence Theory (DST) \cite{fidon2024dempster}, in the K-categorization problem, the model attempts to assign a belief distribution to each category and an overall uncertainty of the entire framework. Thus, for each input, there are K+1 non-negative belief distribution values that sum to 1, as shown in Eq. \ref{eq5.1}.

\begin{equation}\label{eq5.1} u_i+\sum_{j=1}^{K}b_{ij}=1,  \end{equation}
where $u_i$ and $b_{ij}$ denote the overall uncertainty and the probability of the $k^{th}$ class for $i^{th}$ sample, respectively.

For $i^{th}$ input, associate the parameters of the Dirichlet distribution $\alpha=[\alpha_1,\cdots,\alpha_K]$ with uncertainty. Then, uncertainty $u$ is computed as follows:
\begin{equation}\label{eq5.3}
    u_i=\frac{K}{\sum_{j=1}^{K} \alpha_{ij}}=\frac{K}{S_i}.
\end{equation}

In this context, $S_i=\sum_{j=1}^{K}\alpha_{ij}$ is the strength of the Dirichlet distribution, which can be thought of as the total amount of evidence.

For the $i^{th}$ input, predicted probability \( p_{ij} \) for $j^{th}$ category is the mean  of the corresponding Dirichlet distribution and is computed as:
\begin{equation}
    p_{ij} =\frac{\alpha_{ij}}{S_i}.
\end{equation}

For traditional deep neural network-based classifiers, cross-entropy loss is usually used:
\begin{equation}\label{eq5.4}L_{ce}=-\sum_{j=1}^{K}y_{ij}log(p_{ij}), \end{equation}
where $p_{ij}$ is the predicted probability of the $i^{th}$ sample of the $j^{th}$ class. 

For the model in this chapter, the parameters of the Dirichlet distribution $\alpha_i$ can be obtained through the evidential neural network. After a simple modification of Eq. \ref{eq5.4}, the adjusted cross-entropy loss can be obtained, \textit{i.e.},
\begin{displaymath}\label{eq5.51}  L_{ace}=
	\int[\sum_{j=1}^{K}-y_{ij}log(p_{ij})]\frac{1}{B(\alpha_{ij})}\prod\limits_{j=1}^{K}p_{ij}^{\alpha_{ij}-1}dp_i 
  \end{displaymath}
  \begin{equation}\label{eq5.5}  =
	\sum_{j=1}^{K}y_{ij}(\psi(S_i)-\psi(\alpha_{ij})),
  \end{equation}
where $\psi(\cdot)$ denotes the digamma function and the Eq. \ref{eq5.5} is the integral of the cross-entropy loss function determined by $\alpha_i$. 

\begin{table*}[ht]
\centering
\small
\caption{Comparison of \textbf{DUAL} performance with other methods on the DOTA, DIOR, and FGSC-23 test datasets. The table lists the average Top-1 accuracy (\%) for head and tail classes, with ``All'' representing the overall accuracy (\%). The best numbers are highlighted in bold. The backbone for the TLC, BKD, and LAL methods is ResNet32, while ResNet50 is used for the others. Note that we highlight the best performance in \textbf{\textit{bold}} and \underline{underline} the second performance.}
\label{tab:comparison}

\begin{tabular*}{\textwidth}{@{\extracolsep{\fill}}lccccccccc}
\toprule
\multirow{2}{*}{\textbf{Method}} & \multicolumn{3}{c}{\textbf{DOTA}} & \multicolumn{3}{c}{\textbf{DIOR}} & \multicolumn{3}{c}{\textbf{FGSC-23}} \\ 
 \cmidrule(lr){2-4} \cmidrule(lr){5-7} \cmidrule(lr){8-10}
 & \textbf{Head}~($\uparrow$) & \textbf{Tail}~($\uparrow$) & \textbf{All}~($\uparrow$) & \textbf{Head}~($\uparrow$) & \textbf{Tail}~($\uparrow$) & \textbf{All}~($\uparrow$) & \textbf{Head}~($\uparrow$) & \textbf{Tail}~($\uparrow$) & \textbf{All}~($\uparrow$) \\ 
 \midrule
SADE \cite{SADE}  & 94.27 & 88.67 & 93.57 & 88.68 & \underline{86.90} & 88.40 & 68.70 & \underline{76.08} & 70.79 \\
RIDE \cite{RIDE}  & 85.10 & 78.15 & 81.54 & 88.33 & 83.19 & 87.57 & 42.68 & 57.22 & 52.27 \\
ResLT \cite{Reslt} & 94.75 & 81.74 & 94.97 & 78.81 & 81.05 & 72.95 & 64.38 & 63.19 & 62.55 \\
LDMLR \cite{LDMLR} & 87.33 & 80.92 & 92.95 & 80.59 & 79.82 & 86.60 & 52.54 & 51.01 & 51.82 \\ 
\hdashline \noalign{\smallskip}
TLC \cite{TLC}     & 88.25 & 78.99 & 88.24 & 82.27 & 76.39 & 82.70 & 29.00 & 68.50 & 44.12 \\
BKD \cite{BKD}     & 85.16 & 55.96 & 84.74 & 75.31 & 61.43 & 75.71 & 65.41 & 72.05 & 65.94 \\ 
LAL \cite{LAL}     & 93.60 & 68.12 & 92.77 & 82.80 & 76.32 & 84.73 & 54.16 & 49.62 & 53.45 \\
\midrule
T2FTS \cite{zhao2022teaching} & 86.96 & 87.80 & 87.29 & - & - & - & \underline{75.70} & 71.46 & 73.46 \\
EME \cite{bai2024eme}         & 90.32 & \textbf{89.32} & 89.92 & - & - & - & 73.71 & 73.81 & \underline{73.77} \\
TGN \cite{tang2024text}      & \underline{95.56} & 81.49 & \underline{96.10} & \textbf{91.81} & 84.46 & \underline{90.68} & 72.24 & 68.93 & 71.76 \\
\hdashline \noalign{\smallskip}

\textbf{DUAL} & \textbf{97.13} & \underline{89.18} & \textbf{96.66} & \underline{90.54} & \textbf{87.47} & \textbf{91.07} & \textbf{78.21} & \textbf{82.72} & \textbf{79.98} \\
\bottomrule
\end{tabular*}
\end{table*}

Although the loss function described above ensures that the correct labels for each sample produce more evidence than other classes of labels, it does not guaranty that the incorrect labels produce less evidence. Therefore, it is desired that the evidence for incorrect labels in the model be progressively scaled down to close to 0. To this end, the following KL scatter term is introduced:
\begin{displaymath}\label{eq5.61} 
    KL[D(p_{i}|\tilde{\alpha}_{i})||D(p_i|1)]=
	log(\frac{\Gamma(\sum_{j=1}^{K}\tilde{\alpha}_{ij})}{\Gamma(K)\prod_{j=1}^{K}\Gamma(\tilde{\alpha}_{ij})})  
\end{displaymath}
 
 \begin{equation}\label{eq5.6} +\sum_{j=1}^{K}(\tilde{\alpha}_{ij}-1)[\psi(\tilde{\alpha}_{ij})-\psi(\sum_{j=1}^{K}\tilde{\alpha}_{ij})], \end{equation}
where $\tilde{\alpha}_{i}=y_i+(1-y_i)\bigodot\alpha_{i}$ is the Dirichlet distribution-adjusted parameter that avoids the evidence of correct labeling to be zero, and $\Gamma(\cdot)$ is the gamma function.

Thus, given the parameters $\alpha_{i}$ of the Dirichlet distribution for each sample $i$, the loss of specificity for that sample is:

\begin{equation}\label{eq5.7}L_{EDL}=L_{ace}+\lambda_tKL[D(p_i|\tilde{\alpha}_i)||D(p_i|1)],
\end{equation}
where $\lambda_t>0$ is the balancing factor. In the experiment, $\lambda_t$ can be gradually increased as the training progresses to prevent the network from focusing too much on the KL scatter term in the initial stage of training, which may otherwise result in the network not being able to optimize the parameters well enough to output a uniform distribution.

\subsection{Epistemic Uncertainty and Aleatoric Uncertainty}
Predictive Uncertainty (PU) can be decomposed into two parts: EU and AU. EU arises from uncertainty in the model parameters and is typically associated with insufficient training data or knowledge gaps in the model. In contrast, AU reflects the intrinsic noise of the data, which cannot be reduced even if the model achieves perfect fitting. This decomposition is crucial for uncertainty modeling, and PU can be expressed as the sum of EU and AU:

\begin{equation}\text{PU} = \text{EU} + \text{AU}.
\end{equation}

To quantify PU, EU, and AU, predictive distribution entropy measures can be used. The PU of input $x$ can be approximated by the entropy of its predictive distribution $p(y \mid x)$:
\begin{equation}
    \text{PU} = H(p(y \mid x)) = -\sum_{j=1}^{K} p_{ij} \log p_{ij},
\end{equation}
where $H(\cdot)$ denotes the entropy function.

The AU is estimated as the expected entropy over multiple predictions with sampled model parameters $\theta$, \textit{i.e.},
\begin{equation}
\begin{aligned}
\text{AU} &= \mathbb{E}\bigl[H(p(y \mid x, \theta))\bigr] \\
          &= \sum_{j=1}^{K} p_{ij}\bigl[\psi(S_i + 1) - \psi(\alpha_{ij} + 1)\bigr].
\end{aligned}
\end{equation}

The EU can then be computed as the difference between PU and AU:
\begin{equation}
    \text{EU} = H(p(y \mid x)) - \mathbb{E}[ H(p(y \mid x, \theta)) ].
\end{equation}

In the Evidential Deep Learning framework, the Dirichlet distribution parameters $\alpha = [\alpha_1, \alpha_2, \dots, \alpha_K]$ allow the use of $K/S$ as a measure of EU, where $K$ is the number of classes and $S = \sum_{c=1}^K \alpha_c$. Figure \ref{fig:K/S_EU} shows that $K/S$ is highly correlated with entropy-based EU. Compared to entropy, $K/S$ yields a more evenly distributed range of values, making it more suitable for use as a weight in loss functions. Therefore, we choose $K/S$ as the metric for EU to guide uncertainty-aware optimization.

\subsection{Uncertainty-aware long-tailed learning}

To address the challenge of heterogeneous sample learnability in long-tailed remote sensing classification, we utilize EU and AU to optimize the training process through reweighting and label smoothing dynamically. The details of these mechanisms and the final loss function are described below.

\noindent \textbf{Sample Reweighting with EU.} We leverage the EU to dynamically adjust sample weights during training, emphasizing samples with higher EU to strengthen the learning of underrepresented classes. Specifically, for a sample $i$, its training weight $w_i$ is computed as:
\begin{equation}
    w_i = (2\times\text{EU}_i)^{\sigma},
\end{equation}
where $\sigma$ is an exponential scaling factor (typically $\sigma \in [1, 5]$) that amplifies differences in EU, giving larger weights to high-EU samples while down-weighting confident ones. Multiplying by a factor of 2 is to make the EU interval [0, 2], so that it does not tend to 0 after the exponential scaling. This approach encourages the model to prioritize tail samples with high uncertainty.

\noindent \textbf{Dynamic Label Smoothing with AU.} AU captures inherent noise or ambiguity in the data, such as cloud occlusion or mixed land covers in remote sensing images. To mitigate the negative impact of such samples, we introduce an AU-Based dynamic label smoothing mechanism. Traditional label smoothing modifies a one-hot label $y_i$ as:
\begin{equation}
    \tilde{y}_i = (1 - \epsilon) y_i + \frac{\epsilon}{K},
\end{equation}
where $\epsilon$ is a fixed smoothing factor and $K$ is the total number of predefined categories

We extend this by making $\epsilon_i$ adaptive to the AU value of each sample:

\begin{equation}
\tilde{\epsilon}_i = \text{sigmoid}(\text{AU}_i) \cdot \epsilon,
\end{equation}
where the sigmoid function maps AU to $[0, 1]$. Samples with high AU receive a larger smoothing factor, producing a softer label distribution and reducing overfitting risks.

\noindent \textbf{Final Loss Function.} By combining EU-Based reweighting and AU-driven dynamic label smoothing, we design the final loss function as:

\begin{equation}
    L =  w_i \cdot L_{ace}+ \lambda_t KL[D(p_i|\tilde{\alpha}_i)||D(p_i|1)].
\end{equation}

This combined loss encourages the model to focus on tail classes (via EU) and suppress the influence of noisy samples (via AU), improving both performance and robustness for long-tailed remote sensing classification.

%% file: Sec/4_expers.tex
\section{Experiments}
\label{sec:exprs}

\noindent \textbf{Datasets.}~We evaluate introduced framework on three remote sensing benchmarks: DIOR \cite{li2020dior}, DOTA \cite{xia2018dota}, and FGSC-23 \cite{zhang2020fgsc23}, which cover large-scale object detection and fine-grained classification under complex backgrounds. The details for three datasets are described as follows:
\begin{itemize}
    \item DIOR is a large-scale benchmark for optical remote sensing object detection, consisting of 20 categories with 192,465 annotated instances. It is characterized by high inter-class variability and significant intra-class appearance variations.
    \item DOTA contains 2,806 aerial images with categories that largely overlap with DIOR, but with more complex backgrounds and scale variations.
    \item FGSC-23 focuses on fine-grained ship classification with 23 categories, posing a more challenging long-tailed distribution due to the high similarity between subclasses.
\end{itemize}

Following \cite{tang2024text}, we adopt a head–tail partitioning protocol. The class imbalance is quantified by the Imbalance Ratio (IR): $IR = \max(N^c_{real}) / \min(N^c_{real})$, where $N^c_{real}$ is the sample count of class $c$. Detailed statistics and categories are summarized in Table \ref{tab:dataset_stats}.

\noindent \textbf{Implementation Details.}~We use ResNet-50 as the backbone, initialized with ImageNet pre-trained weights. All experiments are conducted using PyTorch 2.1 on an NVIDIA RTX A6000 GPU (48GB) with CUDA 12.2 and cuDNN acceleration. The training is performed for 100 epochs with a batch size of 64, using the Adam optimizer ($\beta_1=0.9, \beta_2=0.999$) and a cosine learning rate decay from $1 \times 10^{-3}$ to $1 \times 10^{-6}$. Weight decay is set to $1 \times 10^{-4}$. Data augmentation strategies include random cropping, horizontal/vertical flipping, and normalization. The hyperparameters of the uncertainty-aware module are empirically set as $\sigma=3$ and $\lambda=0.2$. We adopt overall accuracy, average class accuracy, average head class accuracy, and average tail class accuracy to evaluate classification performance.

\begin{table}[!t]
\centering
\caption{Statistics of the DOTA, DIOR, and FGSC-23 datasets, where the imbalance ratio and scale range represent the imbalance ratio and scale distribution range in the training dataset.}
\resizebox{\linewidth}{!}{
\begin{tabular}{lcccc}
\toprule
\textbf{Dataset} & \textbf{Class Number} & \textbf{Training Samples} & \textbf{Test Samples} & \textbf{Imbalance Ratio} \\
\midrule
DOTA\cite{xia2018dota} & 15 & 98,906 & 28,853 & 86  \\
DIOR\cite{li2020dior} & 20 & 68,025 & 124,440 & 54 \\
FGSC-23\cite{zhang2020fgsc23} & 23 & 3,256 & 825  & 25 \\
\bottomrule
\end{tabular}
}
\label{tab:dataset_stats}
\end{table}

\subsection{Main Results}
We compare the performance of \textbf{DUAL} with state-of-the-art approaches on three remote sensing long-tailed classification datasets: DOTA, DIOR, and FGSC-23. The evaluation metrics include overall accuracy (Top-1 Acc), average accuracy of head classes, and average accuracy of tail classes. The baselines consist of general long-tailed classification methods as well as remote sensing-specific approaches. The experimental results are shown in Table \ref{tab:comparison}.

On the DOTA dataset, our method achieves 97.13\% average accuracy of head classes, 89.18\% average accuracy of tail classes, and 96.66\% overall accuracy, which correspond to improvements of 1.57\%, 7.69\%, and 0.56\% over TGN, respectively. 
On the DIOR dataset, our average accuracy of tail classes reaches 87.47\%, a 3.01\% increase compared to TGN, further indicating that our method effectively enhances tail class learning. 
On the FGSC-23 dataset, our method achieves 78.21\% average accuracy of head classes, 82.72\% average accuracy of tail classes, and 79.98\% overall accuracy, which yield significant improvements of 5.97\%, 13.79\%, and 8.22\% over TGN, respectively, thereby clearly highlighting its superior robustness in various fine-grained scenarios.

\begin{figure}[!t]
    \centering
    \includegraphics[width=0.9\linewidth]{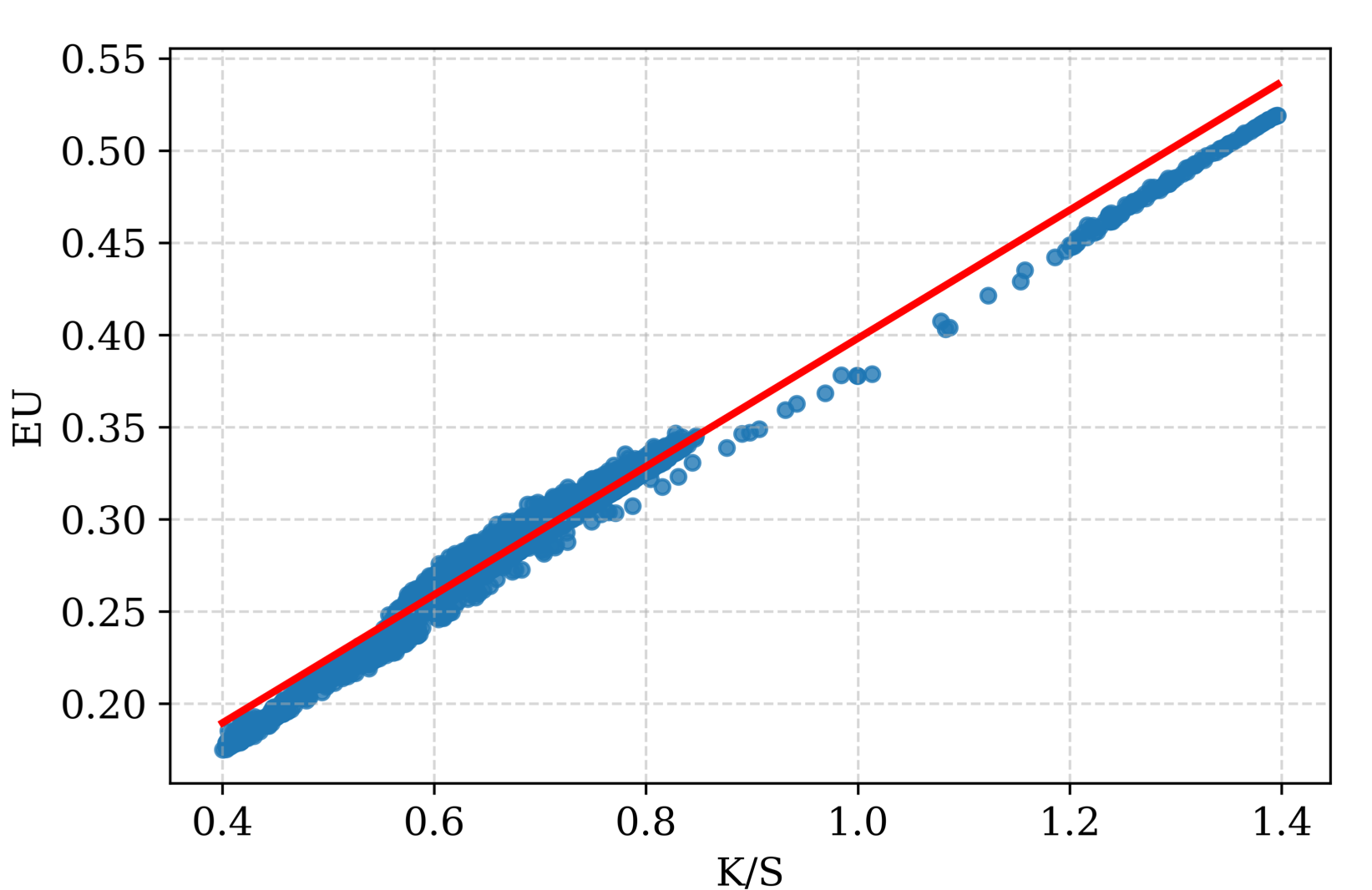}
    \vspace{-5pt}
    \caption{Correlation between the proposed K/S metric and entropy-based EU on FGSC-23 dataset.}
    \label{fig:K/S_EU}
\end{figure}

\begin{table}[!t]
\centering
\caption{Performance (\%) of \textbf{DUAL} with different backbone networks across three remote sensing benchmarks.}
\label{tab:backbone}
\begin{tabular}{lccc}
\toprule
\textbf{Backbone} & \textbf{DOTA} & \textbf{DIOR} & \textbf{FGSC-23} \\
\midrule
EfficientNet-B0 & 96.38 & 91.45 & 77.55 \\
MobileNetV2 & 96.09 & 89.83 & 79.00 \\
ResNet-18 & 95.72 & 89.26 & 78.52 \\
\bottomrule
\end{tabular}
\end{table}

\subsection{Relationship of Different Epistemic Uncertainty}
To validate the core assumption of our method, we analyze the relationship between the proposed K/S metric and entropy-based EU, as well as the correlation between EU and class distribution. Figure \ref{fig:K/S_EU} shows a scatter plot comparing K/S and entropy-based EU on the FGSC-23 dataset, revealing a strong correlation with a Spearman coefficient of 0.99. This indicates that K/S effectively approximates entropy-based EU. However, entropy-based EU values concentrate within a narrow range, limiting sensitivity in dynamically adjusting training weights. In contrast, K/S offers a more uniform distribution over \([0,+\infty]\), making it a better choice for loss weighting to enhance tail class learning.

\begin{table}[!t]
    \centering
    \caption{Performance on FGSC-23 with different exponential scaling factors $\sigma$, while $\epsilon$ is fixed at 0.2.}
    \label{fig:sigma}
    \begin{tabular}{lcc}
        \toprule
        $\sigma$ & \textbf{Acc.(\%)} & \textbf{Avg Acc.(\%)}\\
        \midrule
        1 & 78.64 & 79.65\\
        2 & 78.40 &78.74\\
        3 & \textbf{79.98} & \textbf{81.35}\\
        4 & 78.88 & 79.75\\
        5 & \textbf{79.98} & 79.75\\
        6 & 74.64 & 73.78\\
        \bottomrule
    \end{tabular}
\end{table}

\subsection{Backbone Generalization}
To verify the generalizability of \textbf{DUAL}, we evaluate its performance across three common backbones: EfficientNet-B0, MobileNetV2, and ResNet-18. As shown in Table~\ref{tab:backbone}, our method consistently achieves strong performance across all three backbones on the DOTA, DIOR, and FGSC-23 datasets. This demonstrates that the effectiveness of our approach is not dependent on a specific network architecture and can be flexibly integrated with various backbone models.

\subsection{Hyperparameter Analysis}

\noindent \textbf{Impact of $\sigma$ in EU-Based Reweighting.}~We investigate the impact of the scaling factor $\sigma$ on FGSC-23, with $\epsilon$ fixed at 0.2. 
As shown in Table~\ref{fig:sigma}, performance peaks at $\sigma = 3$, achieving the highest overall accuracy (79.98\%) and average class accuracy (81.35\%). 
A smaller $\sigma$ (\textit{e.g.}, 1) fails to sufficiently emphasize uncertain samples, while an excessively large $\sigma$ (\textit{e.g.}, 6) overly suppresses sample weights, leading to underfitting in head categories.
This suggests that a moderate $\sigma$ effectively balances learning between head and tail classes.

\noindent \textbf{Influence of $\epsilon$ in AU-Based Label Smoothing.}~We further examine the effect of the parameter $\epsilon$, with $\sigma$ fixed at 3. A higher $\epsilon$ increases the smoothing intensity for samples with high aleatoric uncertainty, aiming to reduce overfitting on noisy or ambiguous inputs.
As shown in Table \ref{fig:epsilon}, performance peaks at $\epsilon = 0.2$, with the highest performance. When $\epsilon$ is too small (\textit{e.g.}, 0.1), the smoothing effect is limited, reducing the model’s robustness to noise. Conversely, as $\epsilon$ increases beyond 0.2, both accuracy metrics gradually decline. These results indicate that moderate smoothing improves generalization under data ambiguity without sacrificing discriminative capacity.

\begin{table}[!t]
\centering
\caption{Performance on FGSC-23 with different \(\epsilon\), while $\sigma$ is fixed at 3.}
\label{fig:epsilon}
\begin{tabular}{cccc}
\toprule
\textbf{$\epsilon$}& \textbf{Acc.(\%)} & \textbf{Avg Acc.(\%)}\\
\midrule
0.1 & 77.55 & 79.23 \\
0.2 & \textbf{79.98} & \textbf{81.35} \\
0.3 & 79.49 & 78.59 \\
0.4 & 79.13 & 78.62 \\
\bottomrule
\end{tabular}
\end{table}

\subsection{Ablation Study}
To validate the contribution of each component in our uncertainty-aware long-tailed learning framework, we conduct ablation experiments on the FGSC-23 dataset. Specifically, we integrate Evidential Deep Learning (EDL), EU-Based reweighting, and AU-Based label smoothing progressively, then evaluate overall accuracy and average class accuracy. The detailed results are summarized in Table~\ref{tab:ablation}.

\begin{table}[h]
\centering
\caption{Ablation study. \(\textbf{EDL.}\) represents the Evidential Deep Learning. \(\textbf{EU.}\) represents the EU-Based reweighting, and \textbf{AU.} indicates the AU-Based label smoothing.} 
\label{tab:ablation}
\begin{tabular}{ccccc} 
\toprule
\textbf{EDL.} & \textbf{EU.} & \textbf{AU.} & \textbf{Acc. (\%)} & \textbf{Avg Acc. (\%)}\\ \midrule
\ding{51} & \ding{55} & \ding{55} & 72.33          & 68.24          \\
\ding{51} & \ding{51} & \ding{55} & 76.58          & 75.32          \\
\ding{51} & \ding{51} & \ding{51} & \textbf{79.98} & \textbf{81.35} \\
\bottomrule
\end{tabular}
\end{table}

%% file: Sec/5_conclusions.tex
\section{Conclusion and Discussions}
\label{sec:Conclusions}
In this paper, we introduce \textbf{DUAL} to decompose Epistemic Uncertainty and Aleatoric Uncertainty. By combining EU-Based sample reweighting and AU-driven dynamic label smoothing, our method significantly improves performance in long-tailed remote sensing classification. 
Extensive experiments on DOTA, DIOR, and FGSC-23 demonstrate the effectiveness of our approach, where tail-class performance is notably improved. Ablation studies confirm the necessity of each component in \textbf{DUAL}.

In the future developments, we will explore adaptive hyperparameter scheduling to enhance robustness under dynamic noise scenarios. It aims to promote the broader application in complex real-world environments.